\documentclass[letterpaper, 10 pt, conference]{ieeeconf}
\IEEEoverridecommandlockouts
\usepackage{amsmath,amssymb,amsfonts,bm}
\usepackage{algorithmic}
\usepackage{graphicx}
\usepackage{textcomp}
\usepackage{xcolor}
\usepackage{lipsum}
\usepackage{balance}
\usepackage[style=ieee,dashed=false,isbn=false,url=false,doi=false]{biblatex}
\def\BibTeX{{\rm B\kern-.05em{\sc i\kern-.025em b}\kern-.08em
    T\kern-.1667em\lower.7ex\hbox{E}\kern-.125emX}}

\DeclareSourcemap{
  \maps{
    \map{
      \pertype{article}
      \step[fieldset=language, null]
      \step[fieldset=url, null]
      \step[fieldset=doi, null]
      \step[fieldset=issn, null]
      \step[fieldset=isbn, null]
      \step[fieldset=note, null]
      \step[fieldset=editor, null]
      \step[fieldset=urldate, null]
      \step[fieldset=file, null]
    }
  }
}

\DeclareSourcemap{
  \maps{
    \map{
      \pertype{inproceedings}
      \step[fieldset=language, null]
      \step[fieldset=url, null]
      \step[fieldset=doi, null]
      \step[fieldset=issn, null]
      \step[fieldset=isbn, null]
      \step[fieldset=note, null]
      \step[fieldset=editor, null]
      \step[fieldset=urldate, null]
      \step[fieldset=file, null]
    }
  }
}

\newif\ifblindmode
\blindmodefalse   

\NewDocumentCommand{\blind}{ m m }{%
  \ifblindmode
    \IfBlankTF{#2}{}{#2}
  \else
    #1
  \fi
}

\begin{document}

\title{Autonomous Whole-body\\Loco-manipulation Using a Snake Robot}

\blind{\author{
Adarsh Salagame$^{1\dagger}$, Henry Noyes$^{1\mathrm{\dagger}}$, Haroon Hublikar$^{1}$, Alireza Ramezani$^{1*}$
\thanks{$^{1}$This author is with the Department of Electrical and Computer Engineering, Northeastern University, Boston MA
        {\tt\small salagame.a, noyes.h, hublikar.h a.ramezani@northeastern.edu*}}%
\thanks{$\dagger$ These authors have equal contribution to this work.}
\thanks{$*$Indicates the corresponding author.}
}}{}
\maketitle

\begin{abstract}
Snake robots can leverage their hyper-redundant morphology for both locomotion and manipulation through distributed body contact, but whole-body loco-manipulation is particularly challenging since the entire body serves simultaneously as propulsive mechanism and manipulation tool. Prior work on wheeled, legged, and snake robot platforms has decoupled these functions through dedicated appendages, body segmentation, or grippers. We present closed-loop loco-manipulation with \blind{our snake robot COBRA}{a snake robot} without this distinction, such that any part of the body may make contact. We develop a contact-implicit dynamic model unifying robot–ground, robot–object, and object–ground interactions within a Moreau–Jean time-stepping framework, and formulate a trajectory optimization over the full-order dynamics, guided by a geometric planner and nominal proportional controller, to autonomously push an object to a goal location during sidewinding. We validate the planner and controller on hardware across varied configurations and present simulation results for a simplified trajectory optimization that resolves robot–object and object–ground contacts.
\end{abstract}

\section{Introduction}

Snake robots have emerged as compelling platforms for operation in diverse and unstructured environments. Their hyper-redundant morphology enables locomotion over sandy slopes \cite{marvi_sidewinding_2014}, flat or stepped rigid terrain \cite{gong_conical_2012, tanaka_climbing_2013}, and through complex obstacles such as ladders \cite{takemori_ladder_2018}, ropes \cite{watanabe_principle_2023}, and pipes \cite{rollinson_pipe_2016}. This environmental versatility constitutes the principal advantage of the snake robot form factor, and the ability to interact with objects in the workspace extends their utility beyond pure locomotion into the domain of manipulation.

Loco-manipulation, the simultaneous execution of locomotion and object manipulation, has been explored across a range of robotic morphologies. Legged and wheeled platforms equipped with dedicated manipulators have demonstrated tasks such as door opening and object transport \cite{sleiman_versatile_2023,sleiman_unified_2021, murooka_humanoid_2021, thakar_survey_2022}, where the legs or wheels provide a stable, repositionable base while an arm executes the manipulation objective. Snake robots, however, lack such a clear functional decomposition between locomotion and manipulation subsystems. Instead, they achieve manipulation through alternative strategies: dedicated end-effector grippers positioned via whole-body posture control, whole-body wrapping, or distributed contact along the body \cite{ji_design_2023, elsayed_mobile_2022, salagame_loco-manipulation_2024}. This dual use of the body for both propulsion and environmental interaction introduces a tightly coupled planning problem that, while more challenging to solve, yields highly adaptable behaviors. Whole-body loco-manipulation formulations on legged platforms have further shown that reasoning over this coupling expands the range of achievable tasks, enabling complex manipulation through coordinated leg and manipulator motions \cite{zhu_versatile_2025}.

\begin{figure}
    \centering
    \includegraphics[width=1\linewidth]{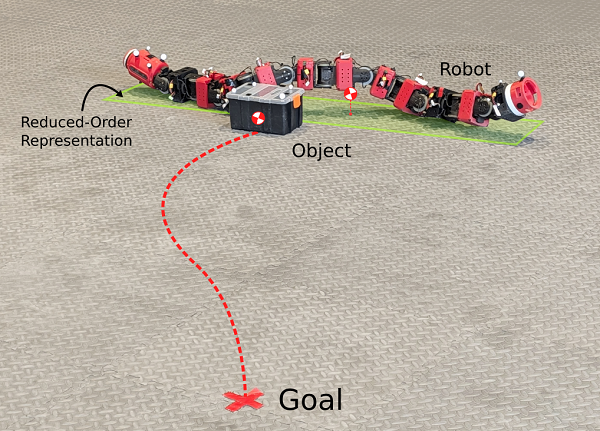}
    \caption{Shows \blind{COBRA}{the robot platform} demonstrating closed-loop loco-manipulation.}
    \label{fig:cover}
\end{figure}
\begin{figure*}
    \centering
    \includegraphics[width=1\linewidth]{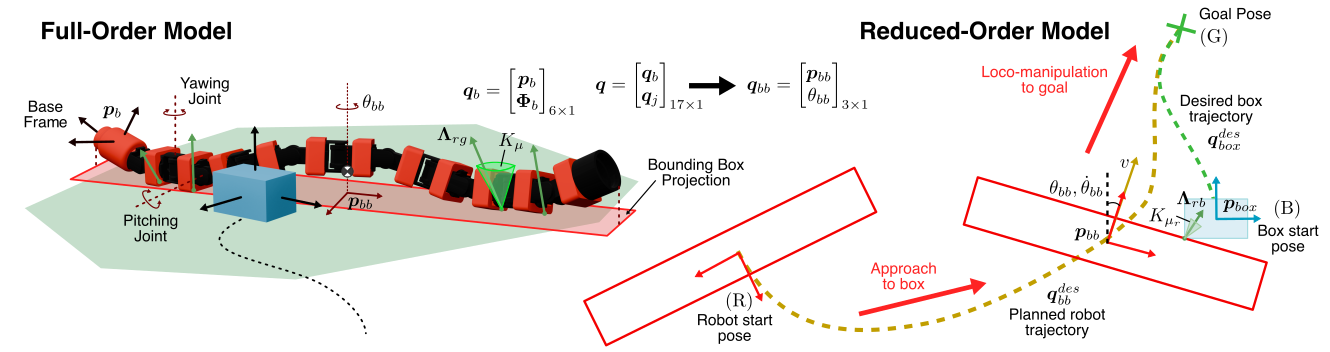}
    \caption{Full-order and reduced-order representations of the snake robot during loco-manipulation. Left: full-order model showing floating-base pose $\bm{q}_b$, joint coordinates $\bm{q}_j$, robot--ground contact impulses $\bm{\Lambda}_{rg}$ with friction cone $K_\mu$, and bounding box projection $\bm{p}_{bb}$. Right: reduced-order planning space with robot start pose (R), box start pose (B), and goal pose (G). The robot reference trajectory $\bm{q}_{bb}^{\mathrm{des}}$ consists of an approach phase and a loco-manipulation phase. The desired box trajectory $\bm{q}_{\mathrm{box}}^{\mathrm{des}}$ defines the box path to the goal. The robot--box contact impulse $\bm{\Lambda}_{rb}$ acts at the contact point subject to friction cone $K_{\mu_r}$.}
    \label{fig:fbd}
\end{figure*}

Applying this perspective to snake robots reveals a richer space of manipulation modalities than is typically exploited. Beyond gripper-based 6-DoF manipulation through bilateral constraints, snake robots can perform non-prehensile single-contact pushing under unilateral constraints as well as whole-body multi-contact interactions for object relocation and reorientation. Despite this breadth, prior loco-manipulation work on snake robots has focused predominantly on gripper-based approaches in which one end of the robot carries a gripper while the remaining body provides locomotion and stability. Ji et al. \cite{ji_design_2023} use the body for locomotion and postural support while a head-mounted gripper lifts objects. Elsayed et al. \cite{elsayed_mobile_2022} partition the robot into a base section, stabilized on a pipe via a helical gait, and a free manipulator section that performs dexterous tasks. Salagame et al. \cite{salagame_non-impulsive_2024} demonstrate both gripper-based pick-and-place relative to the body and whole-body loco-manipulation for subsequent object transport.

In this work, we address whole-body loco-manipulation, tackling the highly coupled motion planning and gait generation problem of pushing an object to a desired goal location. Similar capabilities have been demonstrated in closed-loop on quadrupeds that push objects using their body rather than a dedicated manipulator \cite{jeon_learning_2024, feng_learning_2025}; however, the task is more tightly coupled in snake robots due to the absence of a fixed base. Prior work \cite{salagame_loco-manipulation_2024} demonstrated open-loop whole-body loco-manipulation through pre-defined gait sequences. We build on this foundation by closing the loop with online gait adaptation and motion planning. First, we introduce a geometric planner and feedforward control strategy constructed from gait primitives that provides reliable locomotion and manipulation behaviors. Second, we formulate a contact-implicit trajectory optimization that embeds this feedforward controller as a nominal input, enabling the planner to reason over the contact schedule without prescribing it a priori. We validate the planner and nominal controller through hardware experimentation and validate a simplified trajectory optimization that resolves robot–object and object–ground contacts.

The paper is organized as follows. Section \ref{sec:overview} gives an overview of the robot platform, followed by a discussion of the proposed modeling framework in Section \ref{sec:model}. Section \ref{sec:results} presents results and discussion, followed by concluding remarks in Section \ref{sec:conclusion}.

\section{Problem Statement and Overview of Hardware Platform}
\label{sec:overview}
\blind{COBRA \cite{salagame_crater_nodate, salagame_heading_2024, salagame_loco-manipulation_2024}}{The platform considered in this work \nocite{salagame_crater_nodate, salagame_heading_2024, salagame_loco-manipulation_2024}} (Fig.~\ref{fig:cover}) is a snake-like robot comprising 11 joints arranged in a series kinematic chain with alternating pitch and yaw axes. Each joint is independently actuated by a servo on a shared RS-485 bus, centrally controlled by an Nvidia Jetson Orin NX housed in the head module. A stereo camera and IMU in the head provide localization and environment mapping, while absolute encoders at each joint supply kinematic feedback for the remainder of the body.

We consider the task of whole-body loco-manipulation via sidewinding locomotion: given an initial robot configuration and a known object location, the robot must navigate to the object and push it to a designated goal location using closed-loop control (Fig.~\ref{fig:fbd}). The robot interacts with the object through distributed body contact during sidewinding, without the use of a dedicated end effector or gripper. Robot and object poses are obtained from a motion capture system; object pose estimation and visual-inertial robot localization from the head-mounted stereo camera and IMU is left for future work.

\section{Modeling}
\label{sec:model}


We develop a contact-implicit dynamic model for snake robot loco-manipulation. The robot--ground and robot--box contact interactions are treated within a unified Moreau--Jean time-stepping framework. A reduced-order representation abstracts the oscillatory full-body motion into a stable planning space. These components compose into a single trajectory optimization for loco-manipulation gait generation.

\subsection{Robot Modeling}
We treat the head link as a floating base and define its world-frame pose as $\bm{q}_b = [\bm{p}_b^\top,~\bm{\Phi}_b^\top]^\top$, where $\bm{p}_b, \bm{\Phi}_b \in \mathbb{R}^3$ are the position and Euler-angle orientation of the base, respectively. The generalized position of the robot is $\bm{q} = [\bm{q}_b^\top,~\bm{q}_j^\top]^\top$, where $\bm{q}_j \in \mathbb{R}^{11}$ is the vector of joint angles. The generalized velocity $\dot{\bm{q}} \in \mathbb{R}^{17}$ comprises the base linear velocity $\dot{\bm{p}}_b$, Euler-angle rates $\dot{\bm{\Phi}}_b$, and joint velocities $\dot{\bm{q}}_j$.

Using the Euler-Lagrange formalism, the equations of motion take the form
\begin{equation}
\bm{M}(\bm{q})\,\ddot{\bm{q}}
+ \bm{h}(\bm{q},\dot{\bm{q}})
= \bm{B}\,\bm{\tau}
+ \bm{J}^\top\bm{\lambda},
\label{eq:dyn}
\end{equation}
where $\bm{M}(\bm{q})$ is the mass matrix, $\bm{h}(\bm{q},\dot{\bm{q}})$ encapsulates the Coriolis and gravitational terms, and $\bm{B}$ maps joint torques $\bm{\tau}_j \in \mathbb{R}^{11}$ to generalized coordinates through $\bm{\tau} = [\bm{0}_{6\times1}^\top,~\bm{\tau}_j^\top]^\top$. The vector $\bm{\lambda} \in \mathbb{R}^{3n_c}$ collects the contact forces for $n_c$ contact points, and $\bm{J} = [\bm{J}_1^\top~\cdots~\bm{J}_{n_c}^\top]^\top$, $\bm{J}_i = \partial\bm{p}_c^i / \partial\bm{q}$, is the stacked contact Jacobian, where $\bm{p}_c^i$ denotes the world-frame position of the $i$-th contact point. Next, we build the contact-implicit framework for modeling the interaction of the robot with the ground. 

\subsection{Robot--Ground Interaction Model}
\label{sec:robot-ground}
We partition \eqref{eq:dyn} into actuated and unactuated coordinates:
\begin{equation}
\begin{bmatrix}
\bm{M}_{bb} & \bm{M}_{bj} \\
\bm{M}_{jb} & \bm{M}_{jj}
\end{bmatrix}
\begin{bmatrix}
\ddot{\bm{q}}_b \\ \ddot{\bm{q}}_j
\end{bmatrix}
+
\begin{bmatrix}
\bm{h}_b \\ \bm{h}_j
\end{bmatrix}
=
\begin{bmatrix}
\bm{0} \\ \bm{\tau}_j
\end{bmatrix}
+
\begin{bmatrix}
\bm{J}_b^\top \\ \bm{J}_j^\top
\end{bmatrix}
\bm{\lambda}.
\label{eq:part-dyn}
\end{equation}

We treat joint acceleration $\ddot{\bm{q}}_j$ as the control input. The servos track commanded accelerations via internal PID control with encoder feedback, so any $\ddot{\bm{q}}_j$ within the servo bandwidth is directly achievable. When needed, the actuated row of \eqref{eq:part-dyn} recovers the corresponding joint torques for constraint enforcement.

The $n_c$ contact points on the robot body are partitioned
into $n_{rg}$ robot-ground and $n_{rb}$ robot-box contacts,
with corresponding impulses $\bm{\Lambda} = [\bm{\Lambda}_{rg}^\top,~\bm{\Lambda}_{rb}^\top]^\top$ and Jacobians $\bm{J} = [\bm{J}_{rg}^\top,~\bm{J}_{rb}^\top]^\top$.

The unactuated row yields the floating-base dynamics:
\begin{equation}
\ddot{\bm{q}}_b = \bm{M}_{bb}^{-1}\bigl(
\bm{J}_b^\top\bm{\lambda}
- \bm{h}_b
- \bm{M}_{bj}\ddot{\bm{q}}_j\bigr).
\label{eq:cont-base-dyn}
\end{equation}
The velocity of each contact point on the body depends on both base and joint velocities:
\begin{equation}
\dot{\bm{p}}_c
= \bm{J}_b\,\dot{\bm{q}}_b
+ \bm{J}_j\,\dot{\bm{q}}_j.
\label{eq:contact-vel}
\end{equation}
Using the Moreau--Jean time-stepping scheme \cite{studer_numerics_2009, moreau_unilateral_1988}, we discretize \eqref{eq:cont-base-dyn} at the $k^{\mathrm{th}}$ step with step size $\Delta t$:
\begin{equation}
\dot{\bm{q}}_b^{k+1} = \dot{\bm{q}}_b^k
+ \bm{M}_{bb}^{-1}\bigl(
\bm{J}_b^\top\bm{\lambda}
- \bm{h}_b
- \bm{M}_{bj}\ddot{\bm{q}}_j\bigr)\Delta t.
\label{eq:disc-base}
\end{equation}
Left-multiplying \eqref{eq:disc-base} by $\bm{J}_b$ and adding the known joint contribution from \eqref{eq:contact-vel} yields the Delassus form: 

\begin{equation}
\bm{\gamma}^{k+1}
= \bm{v}_f^k + \bm{G}^k\,\bm{\Lambda}^k,
\label{eq:delassus}
\end{equation}

where $\bm{\gamma}^{k+1} = \bm{J}_b\,\dot{\bm{q}}_b^{k+1} + \bm{J}_j\,\dot{\bm{q}}_j^{k+1}$ is the contact-point velocity at the next step, $\bm{v}_f^k = \bm{J}_b\,\dot{\bm{q}}_b^k + \bm{J}_j\,\dot{\bm{q}}_j^{k+1} + \bm{J}_b\,\bm{M}_{bb}^{-1} (-\bm{h}_b - \bm{M}_{bj}\ddot{\bm{q}}_j)\Delta t$ is the free velocity, $\bm{G}^k = \bm{J}_b\,\bm{M}_{bb}^{-1}\bm{J}_b^\top$ is the Delassus matrix, and $\bm{\Lambda}^k = \bm{\lambda}\,\Delta t$ is the contact impulse. Since $\ddot{\bm{q}}_j$ is prescribed, $\dot{\bm{q}}_j^{k+1}$ is known and enters entirely through $\bm{v}_f^k$.

Contact interactions are subject to unilateral constraints
and Coulomb friction. The feasible set for each
robot-ground contact impulse is the friction cone:
\begin{equation}
K_\mu = \bigl\{
[\bm{\Lambda}_T^\top,\,\Lambda_N]^\top
~\big|~
\|\bm{\Lambda}_T\| \le \mu\,\Lambda_N,~
\Lambda_N \ge 0
\bigr\},
\label{eq:fric-cone-rg}
\end{equation}

where $\bm{\Lambda}_T \in \mathbb{R}^2$ and $\Lambda_N \in \mathbb{R}$ are the tangential and normal components, and $\mu$ is the coefficient of friction. Robot-box contact impulses are subject to the friction cone $K_{\mu_r}$, defined identically with coefficient $\mu_r$. The contact impulses in \eqref{eq:delassus} are obtained by solving the normal-cone inclusion at the velocity level \cite{studer_numerics_2009}:

\begin{equation}
-\bm{\gamma}^{k+1}
\in \mathcal{N}_{K}(\bm{\Lambda}^k),
\label{eq:nci}
\end{equation}

where $\mathcal{N}_{K}(\bm{\Lambda}^k)$ denotes the normal cone to the Cartesian product $K = K_\mu^{n_{rg}} \times K_{\mu_r}^{n_{rb}}$ at $\bm{\Lambda}^k$. For multiple contacts, this is solved via projected Gauss-Seidel iteration with projection onto the appropriate cone at each step.

The base state $[\bm{q}_b^{k+1},\,\dot{\bm{q}}_b^{k+1}]$ is then obtained from \eqref{eq:disc-base} via semi-implicit Euler
integration:
\begin{equation}
\bm{q}_b^{k+1}
= \bm{q}_b^k
+ \dot{\bm{q}}_b^{k+1}\,\Delta t.
\label{eq:base-update}
\end{equation}

The robot--box contact impulse $\bm{\Lambda}_{rb}^k$, resolved within the PGS solve above, acts as an external wrench on the box. We now develop the box dynamics under this wrench and the resulting ground contact response.

\subsection{Robot--Box Interaction Model}
\label{sec:locoman}
\begin{figure}
    \centering
    \includegraphics[width=0.8\linewidth]{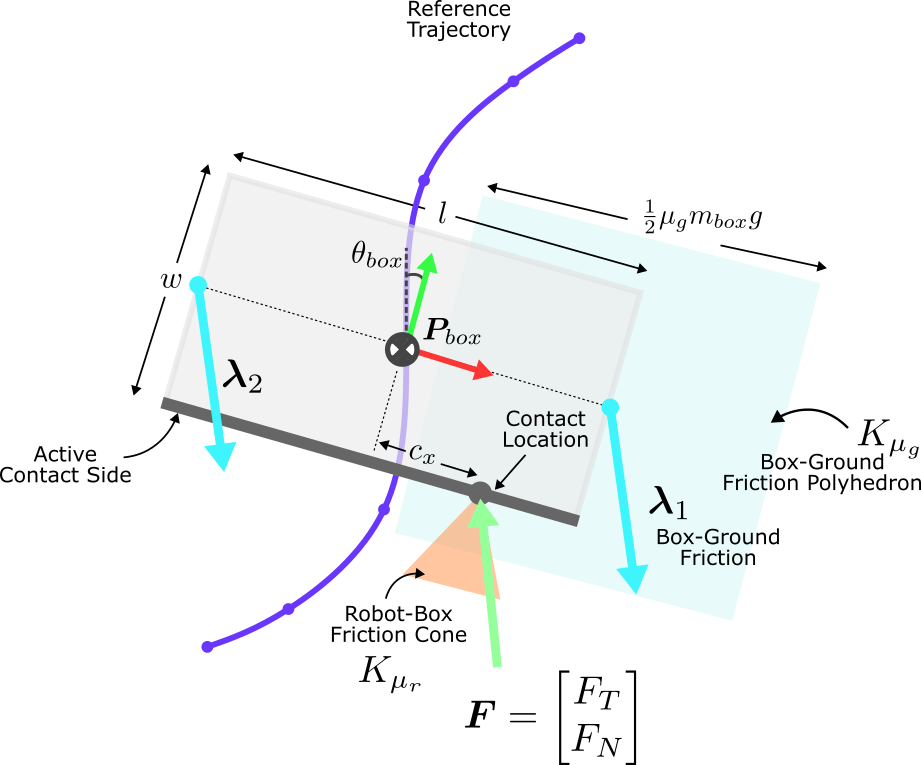}
    \caption{Planar free body diagram of the box. The robot applies force $\bm{F}$ at location $c_x$ on the active contact side. Ground contacts at $\pm l/2$ exert friction forces $\bm{\lambda}_1$, $\bm{\lambda}_2$.}
    \label{fig:box-fbd}
\end{figure}
We model the box as a planar rigid body on the horizontal ground plane with pose $\bm{q}_{\mathrm{box}} = [\bm{p}_{\mathrm{box}}^\top,\;\theta_{\mathrm{box}}]^\top \in \mathbb{R}^3$. The robot contacts the box at a single point on a designated active contact side (Fig.~\ref{fig:box-fbd}) at location $c_x$ along the box $x$-axis measured from the box origin. The contact impulse $\bm{\Lambda}_{rb}^k$ at this point is resolved as part of the PGS solve of \eqref{eq:delassus}--\eqref{eq:nci}, subject to the friction cone $K_{\mu_r}$. The corresponding external wrench on the box is:
\begin{equation}
\bm{W}_{rb}^k
= \begin{bmatrix}
\bm{\Lambda}_{rb}^k \\[2pt]
\bm{r}_{rb} \times \bm{\Lambda}_{rb}^k
\end{bmatrix},
\label{eq:box-wrench}
\end{equation}

where $\bm{r}_{rb}$ is the moment arm from the box center of mass to the contact point, with components determined by $c_x$ and the box half-width $w/2$.

We approximate the ground--box interaction by two point contacts placed symmetrically about the box center of mass at positions $\pm l/2$ along the support region. Since the motion is horizontal, the out-of-plane normal force at each contact is $\tfrac{1}{2}\,m_{\mathrm{box}}\,g$, and each ground contact $i \in \{1,2\}$ exerts an in-plane force $\bm{\lambda}_i \in \mathbb{R}^2$ subject to the friction polyhedron:
\begin{equation}
K_{\mu_g}^i = \bigl\{
\bm{\lambda}_i \in \mathbb{R}^2
~\big|~
\|\bm{\lambda}_i\|_2
\le \tfrac{1}{2}\,\mu_g\,m_{\mathrm{box}}\,g
\bigr\},
\label{eq:box-fric-bound}
\end{equation}

where $\mu_g$ is the ground friction coefficient. The translational and rotational dynamics of the box are:
\begin{align}
m_{\mathrm{box}}\,\ddot{\bm{p}}_{\mathrm{box}}
&= \bm{F} + \bm{\lambda}_1 + \bm{\lambda}_2,
\label{eq:box-trans} \\
I_{\mathrm{box}}\,\ddot{\theta}_{\mathrm{box}}
&= \bm{r}_{rb} \times \bm{F}
+ \bm{r}_1 \times \bm{\lambda}_1
+ \bm{r}_2 \times \bm{\lambda}_2,
\label{eq:box-rot}
\end{align}

where $\bm{F} = \bm{\Lambda}_{rb}^k / \Delta t$ is the contact force, $I_{\mathrm{box}}$ is the moment of inertia about the center of mass, and $\bm{r}_1$, $\bm{r}_2$ are the moment arms to the ground contact points at $\pm l/2$.

In contrast to the robot-ground problem, the fixed contact configuration and low dimensionality of the planar box model
permit the ground contact forces to be recovered without
iterative solvers. Discretizing \eqref{eq:box-trans}--\eqref{eq:box-rot} in impulse form and enforcing zero ground contact-point velocity at the next step yields the ground impulses:
\begin{equation}
\bm{\Lambda}_g^k
= -\bigl(
\bm{J}_g\,\bm{M}_{\mathrm{box}}^{-1}\,\bm{J}_g^\top
\bigr)^{-1}
\bigl(
\bm{J}_g\,\dot{\bm{q}}_{\mathrm{box}}^k
+ \bm{J}_g\,\bm{M}_{\mathrm{box}}^{-1}\,
\bm{W}_{rb}^k
\bigr),
\label{eq:box-ground-impulse}
\end{equation}

where $\bm{M}_{\mathrm{box}} = \mathrm{diag} m_{\mathrm{box}},\,m_{\mathrm{box}},\, I_{\mathrm{box}})$, $\dot{\bm{q}}_{\mathrm{box}}^k$ is the current box velocity, and $\bm{J}_g$ stacks the two ground contact Jacobians. The resulting impulses are projected onto $K_{\mu_g}^i$ to enforce friction feasibility. The box state is then advanced via semi-implicit Euler integration:
\begin{align}
\dot{\bm{q}}_{\mathrm{box}}^{k+1}
&= \dot{\bm{q}}_{\mathrm{box}}^k
+ \bm{M}_{\mathrm{box}}^{-1}
\bigl(\bm{W}_{rb}^k
+ \bm{J}_g^\top\bm{\Lambda}_g^k\bigr),
\label{eq:box-vel-update} \\
\bm{q}_{\mathrm{box}}^{k+1}
&= \bm{q}_{\mathrm{box}}^k
+ \dot{\bm{q}}_{\mathrm{box}}^{k+1}\,\Delta t.
\label{eq:box-pos-update}
\end{align}

The dynamics above propagate the full robot state $\bm{q}^k$ and box state $\bm{q}_{\mathrm{box}}^k$ forward in time. To define a meaningful cost for trajectory optimization, we require a non-oscillatory representation of the robot pose on the ground plane.

\subsection{Reduced-Order Representation}
\label{sec:rom}
The robot locomotes through body oscillations that produce a repeating pattern of link-ground contacts, generating reaction forces that propel the center of mass. The floating base at the head oscillates continuously in both position and orientation, making it a poor choice for planning and control references. We therefore characterize the robot's configuration through its center-of-mass pose and bounding box (Fig.~\ref{fig:fbd}b).

The center-of-mass position is computed from the world-frame positions and masses of all links:
\begin{equation}
\bm{p}_{\mathrm{com}}
= \frac{\sum_{i=0}^{12} m_i\,\bm{p}_i}{\sum_i m_i},
\end{equation}

where $m_i$, $\bm{p}_i$ are the mass and world-frame position of the $i^{\mathrm{th}}$ link. Its orientation is obtained via singular value decomposition of the mass-weighted link positions, which identifies the principal components of inertia in the world frame \blind{\cite{salagame_reduced-order_2025}}{\nocite{salagame_reduced-order_2025}}. The resulting frame is smooth and non-oscillating, providing a stable reference for trajectory planning. The footprint of the robot on the ground is defined as the bounding box of all links.

For planar locomotion on flat ground, the bounding box is fully described by the 3-DoF pose $\bm{q}_{bb} = [x_{bb},~y_{bb},~\theta_{bb}]^\top$, where $\bm{p}_{bb} = [x_{bb},~y_{bb}]^\top$ is the position and $\theta_{bb}$ is the heading angle in the world frame. The map from full configuration to reduced-order coordinates is:
\begin{equation}
\bm{q}_{bb}^k
= \phi(\bm{q}^k),
\label{eq:rom-map}
\end{equation}

where $\phi$ denotes the composition of the center-of-mass
computation and SVD orientation extraction described above. The reduced-order coordinates $\bm{q}_{bb}^k$ and the box pose $\bm{q}_{\mathrm{box}}^k$ together define the loco-manipulation state in the planning space. We now compose the contact-implicit dynamics of Secs.~\ref{sec:robot-ground}--\ref{sec:locoman} with the reduced-order map \eqref{eq:rom-map} into a single trajectory optimization.

\subsection{Contact-Implicit Gait Generation}
\label{sec:gait-gen}

We formulate loco-manipulation as a trajectory optimization over the contact-implicit dynamics developed above. To manage the dimensionality of the 11-joint action space over long horizons, we parameterize the nominal joint motion through a central pattern generator (CPG) following \cite{bing_cpg-based_2017}. A sinusoidal gait with pitch and yaw amplitudes $\alpha_p$, $\alpha_y$ and frequency $\omega$ produces phase-offset oscillations across all joints, with pitching and yawing joints offset by $\pi/2$. The direction of locomotion is controlled via conical sidewinding \cite{gong_conical_2012}: a differential amplitude bias $\Delta$ applied across the yaw joints of each body half produces a net heading rate $\dot{\theta}_{bb}$. This reduces the nominal gait to a single steering parameter $\Delta^k$ at each time step, from which a nominal joint trajectory
$\widetilde{\ddot{\bm{q}}}_j^k$ is obtained by differentiating
the CPG output.

The trajectory optimization refines the joint accelerations
about this nominal. 

\begin{align}
\label{eq:opt}
\min_{\ddot{\bm{q}}_j^{~0:K-1}} \quad
& \sum_{k=1}^{K}
  \|\bm{q}_{\mathrm{box}}^k
  - \bm{q}_{\mathrm{box}}^{k,\mathrm{des}}
  \|^2_{\bm{Q}_{\mathrm{box}}}
  + \|\bm{q}_{bb}^k
  - \bm{q}_{bb}^{k,\mathrm{des}}
  \|^2_{\bm{Q}_{bb}} \\[6pt]
\underset{\forall\, k}{\textrm{s.t.}} \quad
& \eqref{eq:delassus},~\eqref{eq:nci},
  ~\eqref{eq:base-update},
  \label{const:dynamics} \\
& \eqref{eq:box-wrench},
  ~\eqref{eq:box-ground-impulse},
  ~\eqref{eq:box-vel-update},
  ~\eqref{eq:box-pos-update},
  \label{const:box} \\
& \bm{\Lambda}_{rg}^k \in K_\mu^{n_{rg}},\quad
  \bm{\Lambda}_{rb}^k \in K_{\mu_r},\quad
  \bm{\Lambda}_g^k \in K_{\mu_g},
  \label{const:cone} \\
& \bm{q}_{bb}^k = \phi(\bm{q}^k),
  \label{const:bb} \\
& \ddot{\bm{q}}_j^{\min}
  \le \ddot{\bm{q}}_j^k
  \le \ddot{\bm{q}}_j^{\max},
  \label{const:bounds}
\end{align}
where \eqref{const:dynamics} encodes the robot--ground and robot--box contact dynamics; \eqref{const:box} propagates the box state through the contact-implicit box model; \eqref{const:cone} enforces friction feasibility at all three contact interfaces. The desired box trajectory $\bm{q}_{\mathrm{box}}^{k,\mathrm{des}}$ and reduced-order reference $\bm{q}_{bb}^{k,\mathrm{des}}$ are tracked with weights $\bm{Q}_{\mathrm{box}}$ and $\bm{Q}_{bb}$, respectively. The reference $\bm{q}_{bb}^{k,\mathrm{des}}$ guides the robot toward the box when not in contact and maintains contact throughout manipulation. Additional input
regularization terms are omitted for compactness.

The reference trajectory $\bm{q}_{bb}^{k,\mathrm{des}}$ is generated by a nominal planner that switches between approach and manipulation modes, producing geometrically smooth paths between the robot pose and the goal. A proportional controller tracks this reference by modulating $\Delta$, and the resulting
nominal joint accelerations $\widetilde{\ddot{\bm{q}}}_j^k$ serve as both a feedforward input and warm start for the trajectory optimization. The planner and controller are described briefly below.

\subsection{Reference Planner and Nominal Controller}
\label{sec:ref-plan}
\begin{figure}
    \centering
    \includegraphics[width=0.75\linewidth]{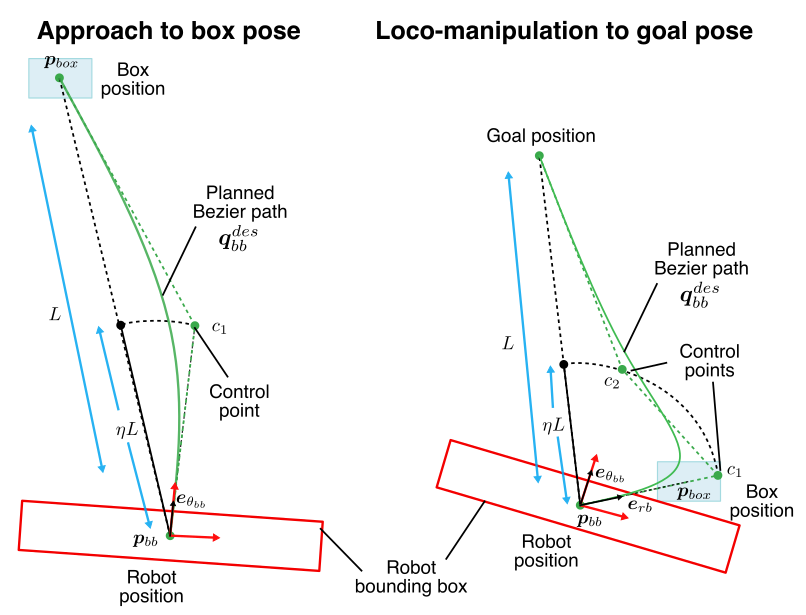}
    \caption{Reference trajectory generation in approach
    mode (left) and loco-manipulation mode (right).
    Approach uses a second-order B\'{e}zier with a single
    control point $\bm{c}_1$ along the heading
    $\bm{e}_{\theta_{bb}}$. Loco-manipulation uses a
    third-order B\'{e}zier with control points $\bm{c}_1$,
    $\bm{c}_2$ along the heading and robot--box directions.
    Both are placed at distance $\eta L$ from their
    respective anchor points.}
    \label{fig:planner}
\end{figure}
The reference trajectory
$\bm{q}_{bb}^{k,\mathrm{des}}$ is generated by a
B\'{e}zier spline whose order depends on the task phase
(Fig.~\ref{fig:planner}).

During approach, the planner constructs a second-order
B\'{e}zier with anchor points at the robot position
$\bm{p}_{bb}$ and the box position
$\bm{p}_{\mathrm{box}}$, and a single interior control
point:
\begin{equation}
\bm{c}_1 = \bm{p}_{bb}
+ \eta\, L\,\bm{e}_{\theta_{bb}},
\end{equation}
where
$L = \|\bm{p}_{\mathrm{box}} - \bm{p}_{bb}\|$ is the
path length, $\eta$ is a proportionality constant, and
$\bm{e}_{\theta_{bb}}$ is the unit vector along the
bounding-box heading. This produces a smooth departure
along the current heading that curves toward the box.

During loco-manipulation, the planner switches to a
third-order B\'{e}zier with anchor points at
$\bm{p}_{bb}$ and the goal position
$\bm{p}_{\mathrm{goal}}$, and two interior control
points:
\begin{align}
\bm{c}_1 &= \bm{p}_{bb}
+ \eta\, L\,\bm{e}_{\theta_{bb}}, \\
\bm{c}_2 &= \bm{p}_{\mathrm{box}}
+ \eta\, L\,\bm{e}_{rb},
\end{align}
where
$L = \|\bm{p}_{\mathrm{goal}} - \bm{p}_{bb}\|$,
and $\bm{e}_{rb}$ is the unit vector from
$\bm{p}_{bb}$ toward $\bm{p}_{\mathrm{box}}$. The
first control point ensures smooth departure along the
heading; the second aligns the path through the box
position along the robot--box approach direction.

A pure-pursuit law generates a desired heading
$\theta_{bb}^{\mathrm{des}}$ by tracking a lookahead
point on the B\'{e}zier path. The heading error
$e_\theta = \theta_{bb}^{\mathrm{des}} - \theta_{bb}$
is regulated by a proportional controller:
\begin{equation}
\Delta = k_\Delta\, e_\theta.
\label{eq:prop-ctrl}
\end{equation}
The forward velocity $v$ is fixed by the gait oscillation
frequency $\omega$ and is not modulated by the controller.
When $\|\bm{p}_{bb} - \bm{p}_{\mathrm{goal}}\|$ falls
below a proximity threshold, the robot arrests forward
motion. If $|e_\theta|$ exceeds a turning threshold, the
gait switches to a turn-in-place mode to realign the
heading before resuming forward locomotion, avoiding the
large-radius arcs inherent to pure pursuit at high heading
errors.

The nominal joint accelerations
$\widetilde{\ddot{\bm{q}}}_j^k$ produced by this controller
through the CPG provide the feedforward input and warm
start for \eqref{eq:opt}.

\section{Results and Discussion}
\label{sec:results}
\begin{figure*}
    \centering
    \includegraphics[width=1\linewidth]{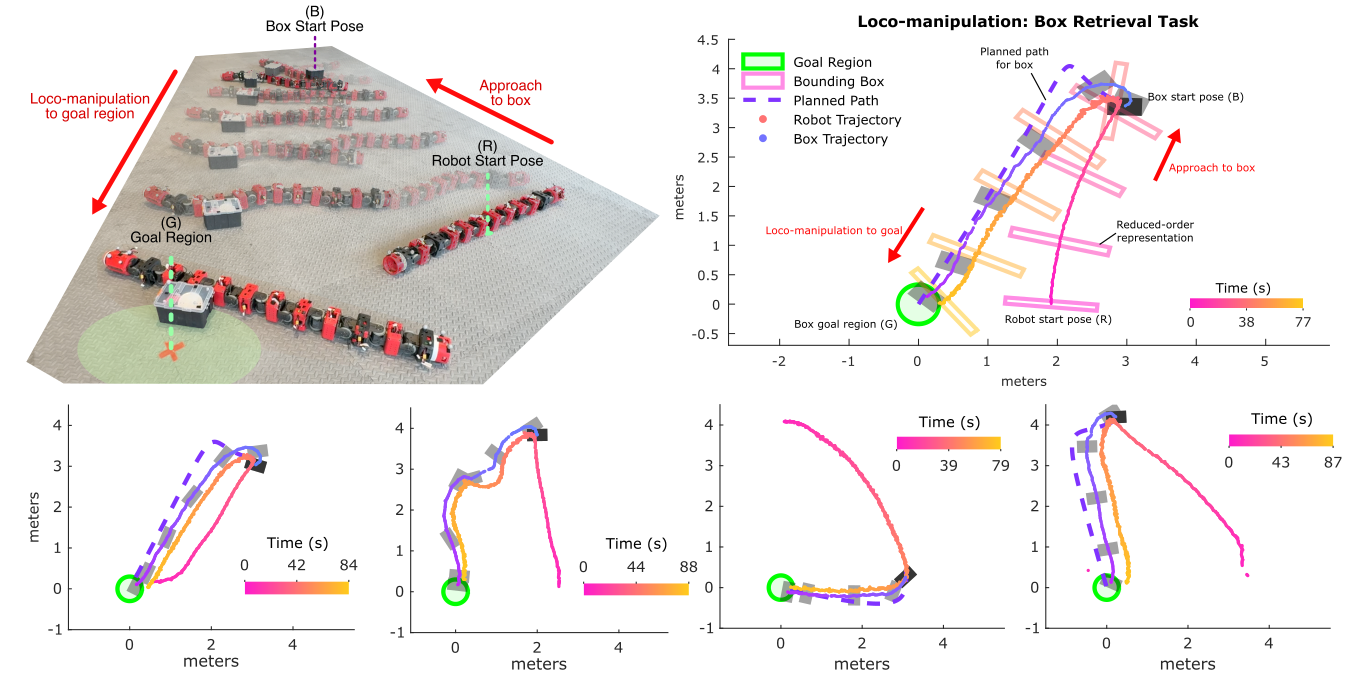}
    \caption{Hardware loco-manipulation results. Top left: composite snapshots of a representative trial showing robot start pose (R), box start pose (B), and goal region (G). Top right: corresponding trajectory plot with reduced-order bounding box, robot center-of-mass trajectory, box trajectory, and planned B\'{e}zier path. Color gradient indicates time progression. Bottom: trajectory plots from four additional trials with varied start and goal configurations.}
    \label{fig:snapshots}
\end{figure*}
\begin{figure}
    \centering
    \includegraphics[width=1\linewidth]{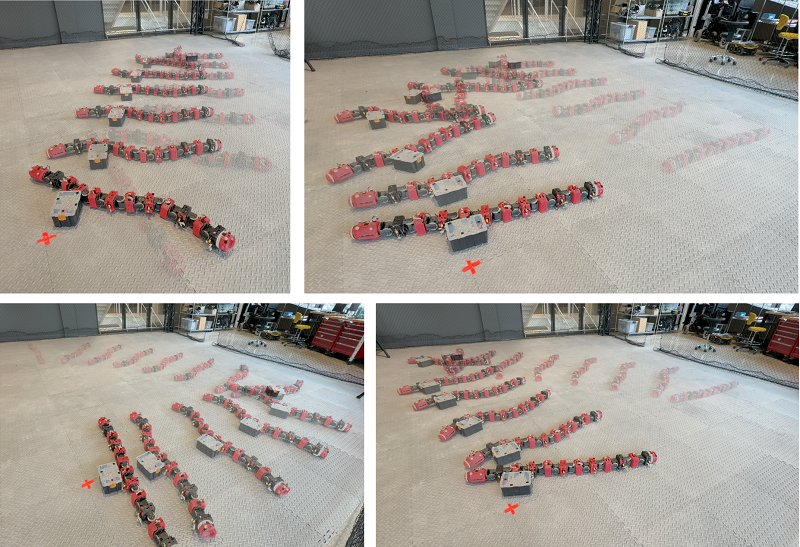}
    \caption{Composite snapshots from four hardware loco-manipulation trials corresponding to the trajectory plots in Fig.~\ref{fig:snapshots} (bottom). The red cross marks the goal position in each trial. Faded overlays show intermediate robot and box poses.}
    \label{fig:snapshots-all}
\end{figure}
\begin{figure*}
    \centering
    \includegraphics[width=1\linewidth]{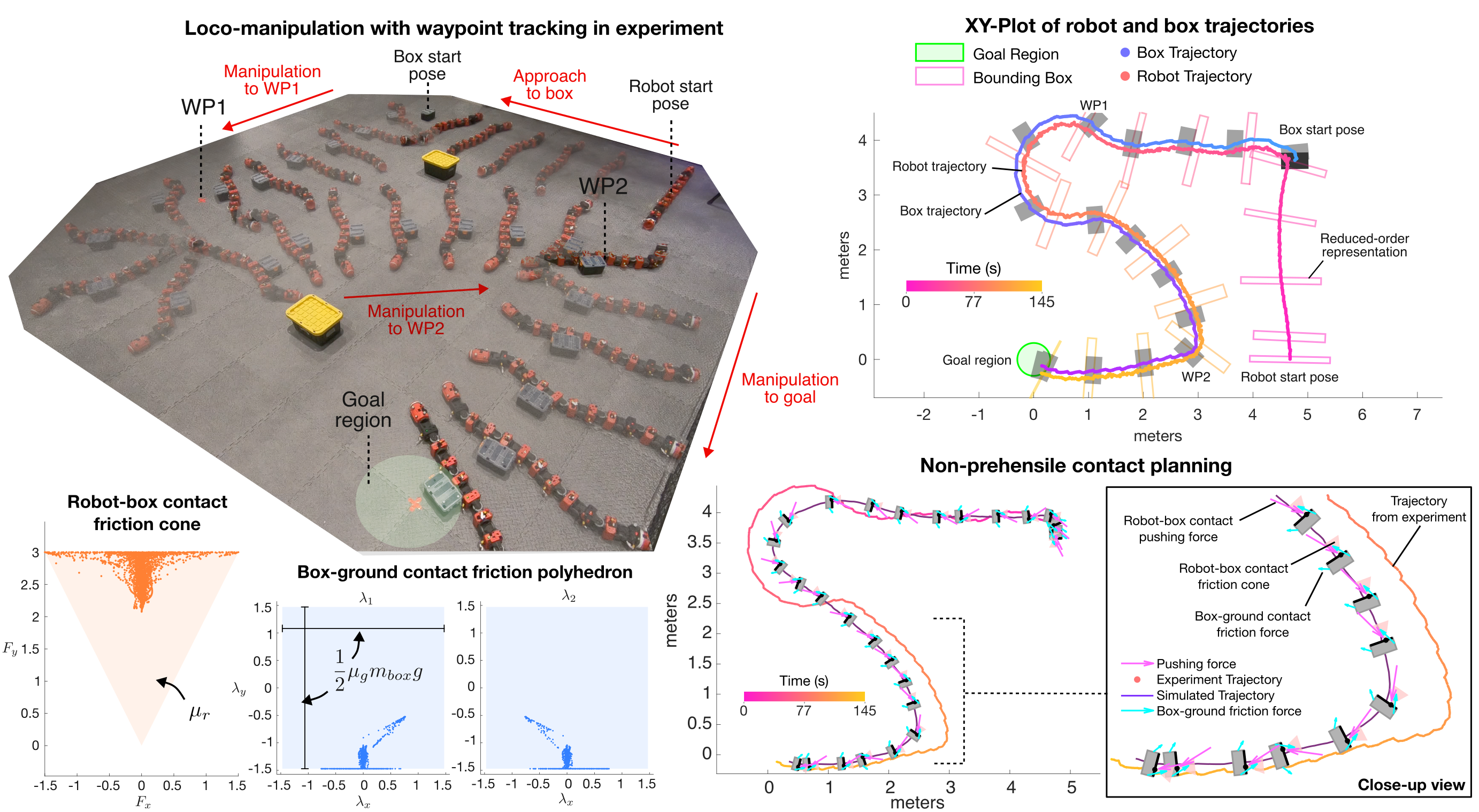}
    \caption{Waypoint tracking experiment and contact-implicit model validation. Top left: composite snapshots showing approach, manipulation through waypoints WP1 and WP2, and delivery to the goal region. Top right: trajectory plot with robot trajectory, box trajectory, and reduced-order bounding box. Bottom left: robot--box contact force distribution against friction cone $K_{\mu_r}$. Bottom center: box--ground friction forces $\bm{\lambda}_1$, $\bm{\lambda}_2$ against friction polyhedron $K_{\mu_g}^i$. Bottom right: box model validation via single shooting---simulated box trajectory overlaid on experimental trajectory with instantaneous pushing forces and ground friction forces at each time step.}
    \label{fig:s-trajectory}
\end{figure*}
\begin{figure}
    \centering
    \includegraphics[width=1\linewidth]{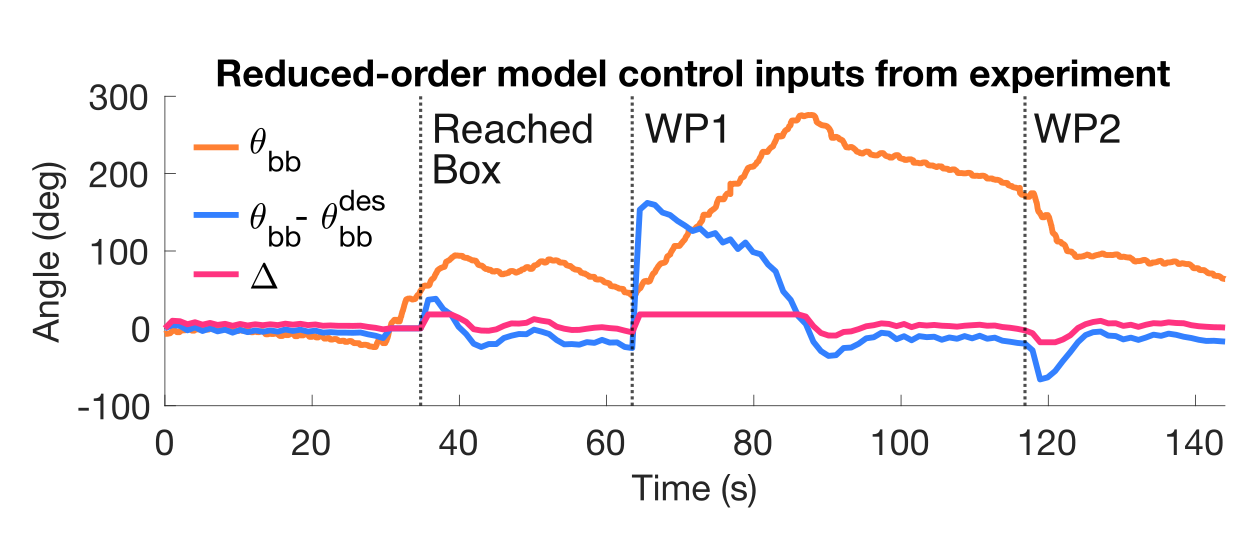}
    \caption{Reduced-order control signals during the waypoint tracking trial of Fig.~\ref{fig:s-trajectory}. Robot heading $\theta_{bb}$, heading error $\theta_{bb} - \theta_{bb}^{\mathrm{des}}$, and steering bias $\Delta$. Vertical dashed lines mark box acquisition and waypoint transitions (WP1, WP2).}
    \label{fig:delta}
\end{figure}


We implement the planner and nominal controller on hardware and test autonomous loco-manipulation performance with various configurations of robot and object poses with respect to the goal. We then test a longer trajectory where the robot must push the box through a set of waypoints before reaching the goal. Finally, we evaluate the box-ground and robot-box contact resolution in simulation using a simplified version of the trajectory optimization presented in \eqref{eq:opt} which omits the robot-ground interaction dynamics and directly solves for \(\bm F,~c_x\). In experiment, robot and object poses are obtained from motion capture. In future work, these will be replaced by visual-inertial localization and object detection using the onboard stereo camera and IMU.

We first discuss the loco-manipulation experiments using the planner and nominal controller. The planner proportionality constant is set to $\eta = 0.4$, the pure-pursuit lookahead distance to $d_{\mathrm{la}} = 0.3$~m, and the controller proportional gain to $k_\Delta = 0.2$. All experiments use a nominal sidewinding gait with $\alpha_p = 15^\circ$, $\alpha_y = 40^\circ$, and frequency $\omega = 0.5$~Hz, producing an average body-frame speed of $v = 0.1~\mathrm{m\,s^{-1}}$. The steering bias $\Delta$ is bounded to $[-15^\circ,\, +15^\circ]$, and the controller runs at $0.75$~Hz. The box used in all experiments has mass $m_{\mathrm{box}} = 0.5$~kg and dimensions $l \times w = 0.2 \times 0.12$~m. 

Fig.~\ref{fig:snapshots} (top left) shows the experimental setup and composite snapshots for a representative trial, with the robot starting at (R), approaching the box at (B), and pushing it to the goal region (G). Fig.~\ref{fig:snapshots} (top right) shows the corresponding trajectory plot, with the reduced-order representation $\bm{q}_{bb}$, the robot center-of-mass trajectory, the box trajectory, and the planned B\'{e}zier path. The bounding box representation tracks the planned path smoothly despite the oscillatory nature of the underlying gait. The box trajectory closely follows its planned path during the manipulation phase, with the deviation between the two remaining within the box footprint throughout.

Fig.~\ref{fig:snapshots} (bottom) presents trajectory plots from four additional trials with varied robot start poses, box positions, and goal locations. Across trials, the approach  phase exhibits smooth curvature consistent with the B\'{e}zier parameterization, and the transition from approach to manipulation occurs without discontinuity in the reduced-order trajectory. The box trajectories show a consistent lateral offset arising from the approach and turning maneuvers, however, the robot accounts for this offset and consistently places the box in the goal region over all experiments.

Over 20 trials spanning a wide spatial distribution of start and goal configurations, the robot achieved a 100\% success rate in transporting the box to within $0.2$~m of the goal pose. Fig.~\ref{fig:snapshots-all} shows snapshots of the corresponding hardware experiments.

Fig.~\ref{fig:s-trajectory} presents a longer loco-manipulation trial in which the robot navigates the box through two intermediate waypoints (WP1, WP2) before reaching the goal. The composite snapshots (top left) show the full sequence: approach to the box, manipulation through WP1 and WP2, and final delivery to the goal region. The corresponding trajectory plot (top right) shows the reduced-order representation $\bm{q}_{bb}$ remains smooth throughout, including during the sharp heading reversals required between waypoints. The box trajectory tracks the planned path through all three segments, with the robot maintaining contact throughout the manipulation phase.

Fig.~\ref{fig:delta} shows the reduced-order control
signals for this trial. The heading $\theta_{bb}$ undergoes
excursions exceeding $200^\circ$ as the robot reorients
between waypoints, while the steering bias $\Delta$ remains
within its $\pm 15^\circ$ bounds throughout. The heading
error $\theta_{bb} - \theta_{bb}^{\mathrm{des}}$ is
regulated to near zero during steady manipulation, with
transient peaks at waypoint transitions where the planner
generates a new B\'{e}zier path.

The box trajectory recorded from this trial serves as a
reference for validating the contact-implicit box model of
Sec.~\ref{sec:locoman}. The robot--box and box--ground friction coefficients used in simulation are $\mu_r = 0.5$ and $\mu_g = 0.5$, respectively. The pushing force $\bm{F}$ and
contact location $c_x$ required to produce this trajectory
are computed by solving a reduced form of \eqref{eq:opt} in
which the robot dynamics are removed and $(\bm{F},\, c_x)$
are the decision variables directly. The problem is solved
via single shooting with a horizon of $1.25$~s and time step
$\Delta t = 0.025$~s.

Fig.~\ref{fig:s-trajectory} (bottom right) shows the
result. The simulated system successfully recovers a
force sequence that drives the box along the experimental
trajectory. The close-up view shows the instantaneous
pushing and ground friction forces acting on the box at
each time step. Fig.~\ref{fig:s-trajectory} (bottom left)
shows the force distributions and cone feasibility across
all simulated time steps. The robot--box contact forces
cluster near the normal direction of the active contact
face, consistent with a pushing interaction. The
box--ground friction forces for both $\bm{\lambda}_1$ and
$\bm{\lambda}_2$ (bottom center) concentrate at the
boundary of the friction polyhedron $K_{\mu_g}^i$, indicating
sustained sliding contact consistent with the observed box
motion. The single-shooting solve executes faster than real
time, with the full ${\sim}145$~sec trajectory solved in
under one minute.

The feedforward controller demonstrates reliable and repeatable performance across the tested configurations, achieving consistent object displacement with low variance despite operating without contact feedback. That this level of precision is attainable under open-loop gait execution on flat, rigid ground suggests that the nominal controller provides a strong initialization for optimization-based approaches, reducing the burden on the contact-implicit planner to deviate significantly from the feedforward trajectory. The box-pushing simulation corroborates this: even in the simplified formulation that omits the full robot dynamics, the optimizer recovers dynamically feasible pushing trajectories that respect friction cone constraints at both the robot–object and object–ground interfaces, and does so at computation times compatible with real-time deployment. While non-prehensile pushing has been extensively studied in fixed-base manipulation, the formulation presented here embeds it within a locomotion context in which no dedicated end-effector exists and propulsion and manipulation share the same contact surfaces. The present results validate the individual subsystems: feedforward sidewinding, geometric planning, and contact-implicit object interaction. The next step is to implement full-order whole-body loco-manipulation, in which the optimizer jointly resolves gait generation, terrain adaptation, and object transport within a single contact-implicit trajectory optimization.

\section{concluding remarks}
\label{sec:conclusion}
The contact-implicit modeling framework developed in this work unifies robot–ground, robot–object, and object–ground interactions within a single Moreau–Jean time-stepping formulation, enabling whole-body loco-manipulation without body segmentation or dedicated grippers. A reduced-order representation of the sidewinding gait eliminates the oscillatory dynamics from the planning space, yielding a stable manifold over which both geometric planning and feedback control operate directly.

Hardware experiments demonstrate that this reduced-order planning approach, paired with a nominal proportional controller, is sufficient for reliable object transport: across 20 trials spanning a wide distribution of start and goal configurations, including multi-waypoint trajectories, the system achieved a 100\% success rate in delivering a box to within 0.2~m of the goal. The contact-implicit trajectory optimization, evaluated independently on the object interaction subsystem, recovers physically consistent pushing forces that satisfy all friction bounds and executes faster than real time, confirming its viability as an online planning component.

These results establish the computational and experimental foundations for full-order contact-implicit loco-manipulation, in which gait generation and object interaction are resolved jointly rather than through the decoupled subsystems validated here. Immediate extensions include closing the loop around the full trajectory optimization to enable gait adaptation on uneven terrain, and replacing motion capture with onboard visual-inertial localization and object detection for field deployment.

\section{ACKNOWLEDGMENT}
This work is supported by the Technology Innovation Institute (TII), UAE, and the U.S. National Science Foundation (NSF) CAREER Award (Award No. 2340278). This work is also supported by a NASA Space Technology Graduate Research Opportunity (Grant No. 80NSSC24K1393).

\nocite{krishnamurthy_contact-implicit_2026, krishnamurthy_enabling_2024,wang_quadratic_2025, pitroda_quadratic_2024,salagame_dynamic_2024,salagame_heading_2024,salagame_reduced-order_2025,jiang_hierarchical_2024,jiang_snake_2024,krishnamurthy_thruster-assisted_2024,krishnamurthy_narrow-path_2024,pitroda_conjugate_2025,sihite_posture_2024}

\balance
\printbibliography
\end{document}